\documentclass{llncs}

\usepackage{amsmath}
\usepackage{amssymb}

\usepackage{graphicx}

\title{A Multi-Agent System Approach to Load-Balancing and Resource Allocation for Distributed Computing}

\author{Soumya Banerjee\inst{1,2,3} \and Joshua P. Hecker\inst{4}}

\institute{The Broad Institute of MIT and Harvard, Cambridge, USA
\and Complex Biological Systems Alliance, North Andover, USA
\and Ronin Institute, Montclair, USA
\and Department of Computer Science, University of New Mexico, Albuquerque, USA
\email{neel.soumya@gmail.com}\\
\email{jhecker@cs.unm.edu}}

\date{}

\begin{document}

\maketitle

\section{Abstract}
In this research we use a decentralized computing approach to allocate and schedule tasks on a massively distributed grid. Using emergent properties of multi-agent systems, the algorithm dynamically creates and dissociates clusters to serve the changing resource demands of a global task queue. The algorithm is compared to a standard first-in first-out (FIFO) scheduling algorithm. Experiments done on a simulator show that the distributed resource allocation protocol (dRAP) algorithm outperforms the FIFO scheduling algorithm on time to empty queue, average waiting time and CPU utilization. Such a decentralized computing approach holds promise for massively distributed processing scenarios like SETI@home and Google MapReduce.

\section{Introduction}
Recent years have seen a trend in moving large computational tasks to collections of inexpensive, commercial off-the-shelf (COTS) computers that are geographically distributed. This has contributed significantly to the advancement of science by providing access to large-scale shared computing resources on which to solve computationally expensive problems. Some common examples are SETI@home \cite{anderson2002seti} which runs tasks on millions of computers worldwide and Google MapReduce \cite{dean2004mapreduce} which distributes calculation of web crawled metrics among thousands of computers. This move towards distributed computing has created a need for efficient task allocation and scheduling algorithms. Such algorithms should be very scalable since these systems typically have thousands to millions of computers. They should also be robust to single-point failures and be adaptive to task demand. Recent research on grid resource allocation has focused on volunteer resource allocation, agreement-based resource allocation and economic resource allocation \cite{krawczyk2008grid}. Multi-agent decentralized systems offer an exciting approach to distributed resource allocation. They have emergent global properties which arise from local interactions and have been previously used to model biological phenomena \cite{jacob2004immunity,banerjee_phd,swarm2010,Banerjee2009,Banerjee2011,Banerjee2009a,banerjee2010modular,alife_radar} and solve real-world problems \cite{hecker2008team,Hecker2012,Hecker2013,Hecker2013a,Hecker2015}. Here we use such a decentralized computing approach to allocate and schedule tasks on a grid. The remainder of the paper is organized as follows: Section 3 formalizes the problems and states the assumptions, Section 4 briefly reviews decentralized computing and the advantages it can afford to a distributed allocation problem, Section 5 introduces multi-agent systems, Section 6 introduces the simulator used for the experiments in this paper, Section 7 discusses the dRAP algorithm, Section 8 deals with analysis of the cost of searching through the global queue, Section 9 discusses some dRAP optimization techniques influenced by the immune system, Section 10 deals with experiments and results, Section 11 discusses related work in this area and Section 12 presents concluding remarks and outlines future work.

\section{Statement of Problem and Assumptions}

Assume there is a queue $Q$ of processes waiting to be allocated to processors. Each process is required to declare \textit{a priori} its resource requirements viz. the number of threads into which it can be parallelized ($TH_n$) and the number of system resources it requires (the number of CPUs is assumed to be equal to the number of threads which can be run in parallel, $CPU_{req}$). Our system departs from traditional resource allocation techniques in that there is no centralized dispatcher. Instead, we dynamically organize a system of geographically distributed computers into clusters to service each process in $Q$. Over time, clusters of computers are dynamically created, dissociated and created again in order to serve the resource requirements of the processes in $Q$. We define a \textit{cluster} as a network of computers which together can completely service the resource requirements of a single process. Clusters of computers are created so as to be proximal to each other in order to reduce latency and communication costs.

We acknowledge the following assumptions in our system:
\begin{enumerate}
\item Distributed computers can communicate with each other. 
\item There are advantages to computing with geographically proximal computers due to network latency and bandwidth limitations.
\item A new process $P_1$ that comes in the system will declare \textit{a priori} the number of threads that it can be parallelized into and its resource requirements (e.g. the number of CPUs it will require, I/O devices required, amount of memory, etc).
\item The approach will become viable in the asymptotic region of millions or billions of geographically dispersed computers, when there will be expected benefits from a decentralized computing approach that exploits geographical proximity and reduces latency costs, as opposed to a centralized monitor.
\end{enumerate}

\section{Decentralized Computing}
The extreme size of the computing grid and an ever-increasing demand for computational power places exacting demands on any scheduling, allocation and load-balancing algorithm. Here we argue that a decentralized computing paradigm presents an ideal solution to the bottlenecks and single-point failures inherent to a centralized monitor tasked with allocating resources and balancing loads in the grid:

\begin{enumerate}
\item The workload assigned to a centralized monitor increases as computers are added to the computing grid. A decentralized approach can alleviate the computing load on monitors. In this approach, each individual computer, or cluster of computers, will do some computation.
\item A centralized monitor makes the system susceptible to single-point failures. Distributing load balancing and resource allocation tasks to individual computers will increase system robustness.
\item Individual computing nodes are naturally aware of their own workloads. As a result, the decentralized paradigm can achieve application-level resource management with significantly less communication overhead than a centralized monitor.
\item A decentralized system uses peer-to-peer networking to scale communication as the system grows, whereas a centralized monitor has to communicate with an increasing number nodes.
\item A decentralized system is more robust to single node disruptions and failures, whether malicious or benign.
\item A decentralized system may be able to better respond to fluctuations in process requirements e.g. in a scenario where the scheduler has to ``forget" past process requirements and completely rebuild new clusters after servicing one process i.e. there is no locality in process requirements. 
\end{enumerate}

\section{Multi-Agent Systems}
Multi-agent systems use distributed agents to either model or solve a problem. An agent is an entity which matches some real-world object. It could be a biological cell, a virus particle, an ant or in our case an individual computer. A computer program encodes simple rules or behaviors for interacting with other agents. The agents move about in space and interact with other agents in their neighborhood according to the encoded rules. Thus the behavior of low-level entities is specified and high-level behaviors evolve as simulation time progresses. Multi-agent systems emphasize local interactions based on first principles, and these interactions give rise to the complex high-level \textit{emergent properties} of interest. Such systems have been used to model biological phenomenon such as the human immune system \cite{jacob2004immunity}, as well as solve real-world problems like communication between distributed radar transmitters \cite{hecker2008team} and efficient resource collection in swarms of foraging robots \cite{Hecker2012,Hecker2013,Hecker2013a,Hecker2015}.


There is no centralized dispatcher to facilitate the formation and dissociation of clusters in the proposed dRAP algorithm. Instead, the algorithm relies on the \textit{self-emergent} properties of a multi-agent system. A \textit{multi-agent} or \textit{agent-based system} is an architecture in which the global properties of the system emerge from local interactions.
	
The concept of a decentralized system presents a powerful counterpoint to the more common centralized control model often seen in business, government, and military organizations. Decentralization provides a number of important advantages over closed systems, such as robustness, adaptability, flexibility, innovation, and distributed intelligence. The key to this compelling architecture is the impressive ability of a decentralized system to react, mutate, or grow in response to challenging situations.

In any such decentralized system, the agent represents the base unit of computing power for the system. It behaves according to very simple rules. At each unit of model time (or \textit{time step}), the agent senses its immediate local environment and takes actions based on its encoded rules. One rule might instruct the agent to divide if the number of neighbors is greater than 3, while another would cause it to die and be removed from the simulation if the number of neighbors is less than 2. These two examples are rules in the ``Game of Life'' \cite{gardner1970mathematical}, a paradigmatic system where complex patterns arise from local interactions and simple rules. 

If we recast each agent's local sensing functionality as a peer-to-peer communication protocol with other nearby agents, then we can define a new set of rules for each agent that induce actions based on the state of these other, neighboring agents. Using this localized communication scheme, such rule-action pairs can be viewed as instructions for individual agents that produce \textit{decentralized computation} across the system. There is no centralized monitor and yet this system is capable of performing complex computations. In fact, the computational power of such a system of distributed agents acting on simple rules has been proven to be Turing-complete \cite{cook2004universality}.

\begin{figure}
\begin{minipage}[c]{0.5\linewidth}
\centering
\includegraphics[scale=.45]{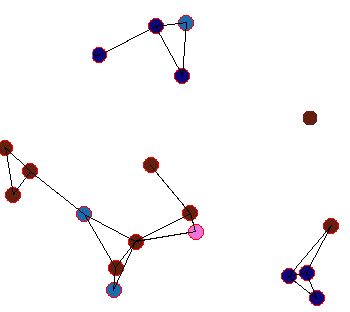}
\caption{\small Agents in large clusters; 1 free agent}
\label{fig:clusters1}
\end{minipage}
\hspace{0.5cm}
\begin{minipage}[c]{0.5\linewidth}
\centering
\includegraphics[scale=.45]{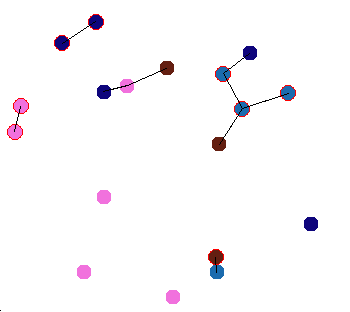}
\caption{Agents in several smaller clusters}
\label{fig:clusters2}
\end{minipage}
\end{figure}

We use such an agent-based system to dynamically create and dissociate clusters based on the resource requirements of each process. A snapshot of this system is shown in Figures \ref{fig:clusters1} and \ref{fig:clusters2}.

\vspace{5 mm}

\vspace{5 mm}

\section{Software Platform}

For this project we utilize the multi-agent simulation toolkit MASON \cite{luke2004mason}. MASON consists of a fairly small and portable set of Java library files that provide for design of both model (the ``algorithm'' component) and visualization (the ``graphical user interface'' component).

The agent, the base component of computation in MASON (as in any multi-agent system), is coded in the familiar object-oriented programming format: the class ``Agent'' that contains all generalized methods and parameters needed for the object ``agent'' that is simply an instantiation of the Agent class. Following this format, each instantiated agent may contain a unique set of parameters, thereby allowing for minor variation in the replicated objects.

Agents are allowed to make decisions (and even communicate with one another) in a randomized batch lock-step. That is, the MASON scheduler moves through the (randomized) queue of all agents at each time step of the simulation. Scheduling of agents continues as long as the simulation itself is running, although the user may interrupt at any point by pausing or stopping the model.

MASON in particular was selected because of its all-in-one toolkit approach, making multi-agent simulation much easier than if done from scratch, as well as the authors' familiarity and experience with the MASON system.

\section{dRAP Algorithm}
The distributed resource allocation protocol (dRAP) is described below and some intended optimizations are suggested for future work. An agent in our system is simply a computer. Each agent has a vector containing the time remaining to finish executing its current process ($time_{rem}$) and the number of CPUs in its current cluster ($CPU_{cluster}$). Each agent (or node) is guaranteed to be in exactly 1 of 4 modes (or states) during the simulation:

\begin{description}
\item[Mode 1:] An agent/node that is currently not part of a cluster and has no task assigned to it
	\begin{enumerate}
	\item The agent scans the queue $Q$, considers the resource requirements $CPU_{req}$ of unallocated tasks, and takes on the task which minimizes the equation $|CPU_{req} - 1|$.
	\end{enumerate}

\item[Mode 2:] An agent/node that is currently not part of a cluster and has a task assigned to it
	\begin{enumerate}
	\item The agent continues executing the task and updates its information vector ($time_{rem}$, $CPU_{cluster}$). 
	\item If the task requirements are not completely satisfied (i.e. if $CPU_{req} > 1$), the agent will query its neighbors and attempt to form a cluster such that $CPU_{req} = CPU_{cluster}$
	\item When the agent finishes executing the task, it returns to \textbf{Mode 1}.
	\end{enumerate}

\item[Mode 3:] An agent/node that is currently part of a cluster and has no task assigned to it
	\begin{enumerate}
	\item The agent scans the queue $Q$, considers the unallocated tasks, and takes on the task which minimizes the equation $|CPU_{req} - CPU_{cluster}|$
	\end{enumerate}

\item[Mode 4:] An agent/node that is currently part of a cluster and has a task assigned to it
	\begin{enumerate}
	\item The agent continues executing the task and updates its information vector ($time_{rem}$, $CPU_{cluster}$)
	\item When the task completes, the agent dissociates from the cluster and returns to \textbf{Mode 1}.
	\end{enumerate}	
\end{description}

A key feature of our algorithm is that nodes query their neighbors (other nodes that are close to them physically) in order to form clusters. This has the effect of reducing latency and communication costs. One optimization to consider would be to delay cluster dissociation in \textbf{Mode 4}. This would lead to \textit{learning} or \textit{memory} in the system where the scheduler would be able to remember past process requirements.

\section{Analysis of Queue Cost}
The dRAP algorithm requires a traversal through the global task queue in \textbf{Mode 1} and \textbf{Mode 3}. The algorithmic complexity is given by $\sum{(n-i)m} = O(n^{2}m)$ where $m$ = the number of tasks in the global task queue, and $n$ = the average number of clusters. At a given timestep, the worst case can be approximated as $O(nm)$.

\section{Optimizations Inspired by the Immune System}
The immune system is able to find rare spatially localized pathogens and eliminate them in a timely manner \cite{swarm2010,banerjee2010modular}. Similar to how in our system clusters of computers find processes, the immune system uses specialized cells to find pathogens in anatomical regions called lymph nodes. In previous work we showed how a sob-modular arrangement of lymph nodes could lead to fast elimination of pathogens in the immune system and also faster search for solutions in immune inspired distributed systems of computers \cite{swarm2010,banerjee2010modular,alife_radar}. Let an artificial lymph node be composed of a number of clusters and a process queue. Also let there be a number of such artificial lymph nodes that have the capability of communicating with each other. An `artificial lymph node` is supposed to be a computer in charge of a number of clusters. This computer will store the process queue and also will have some memory and CPU to communicate with other `lymph nodes`
   
   We are interested in making the system sub-modular so that we can minimize the total time to find a cluster. There is a tradeoff between the local cost and the global cost; 
The local cost is $O(n^{2})$ and the global cost $O(N/n)$.
	The total cost of traversing through the queue in a lymph node and the cost of communicating with other lymph nodes can be summed up as :
	\begin{equation}
	\label{eqn_localglobal}
	t_{total} = t_{local} + t_{global}
	\end{equation}

	\begin{equation}
	\label{eqn_localglobal2}	
	t_{total} = O(n^{2}) + O(N/n)
	\end{equation}
	where where $n$ is the number of clusters in a single lymph node and $N$ is the total number of clusters in the complete system. We assume that the global cost of finding another cluster in another lymph node that can service some process requirement is proportional to the number of lymph nodes (where $N/n$ is the number of lymph nodes in the system).

	Minimizing the total time cost, we get
	$2n - N/n^2 = 0$
	
	\begin{equation}
	\label{eqn_finalscaling}
	n = O(N^{1/3})
	\end{equation}

	This implies that in larger systems (more computers. more clusters and more lymph nodes), the number of clusters within a single lymph node should grow larger but only sublinearly in the number of total clusters in the system. This would balance local costs of queue traversal and global costs of finding another lymph node with another cluster that can service the process.
The key point here is that the number of clusters in a lymph node should scale sub-linearly with the size of the whole system, i.e. if a system of networked artificial lymph nodes were to grow a 1000 times bigger (1000 times more clusters), then the number of clusters within a lymph node need only increase by a factor of 10. Such sub-modular systems inspired by the immune system have been proposed previously for mobile ad-hoc networks, control of mobile robots, intrusion detection systems and peer-to-peer networks \cite{swarm2010,banerjee2010modular,alife_radar}.

	More generally, if the local and global communication costs scale with exponents $\alpha$ and $\beta$, we have
	\begin{equation}
	\label{eqn_localglobal_general}	
	t_{total} = O(n^{\alpha}) + O(N^{\gamma}/n^{\beta})
	\end{equation}
	
Minimizing the expression with respect to $N$, we get	

	\begin{equation}
	\label{eqn_finalscaling_general}
	n = O(N^{ \frac{\gamma}{\alpha + \beta}}) 
	\end{equation}

\begin{enumerate}

\item If $\gamma < \alpha + \beta$ we have sub-linear scaling.
\item If $\gamma > \alpha + \beta$ we have super-linear scaling.
\item If $\gamma = \alpha + \beta$ we have linear scaling.
\item If $\gamma / (\alpha + \beta)$ = 0 we have no scaling (constant).
\item If $\gamma / (\alpha + \beta) < 0$  we have negative scaling.

\end{enumerate}


\section{Experiments}

We conduct several experiments that compare our dRAP algorithm to a null model, i.e. a first-in first-out (FIFO) scheduling system. Additionally, we measure the effective computational complexity of queue traversals 
and examine the scaling properties of our system by varying the number of nodes and measuring the effect on performance. We define two timing metrics on which our system performance will be judged: $T_{complete}$ is the time required to complete all tasks in the queue, and $T_{wait}$ is the average wait time for a task added to the queue. Unless otherwise noted, system parameters are defined as such: number of nodes = 100, number of tasks = 1000, tasks are randomly selected from a normal distribution s.t. $CPU_{req}$ varies from 1 to 5, with initial $time_{rem}$ varying from 25 to 125 in increments of 25. That is, a task $t_i$ with $CPU_{req} = 1$ has an initial $time_{rem} = 25$, and a task $t_j$ with $CPU_{req} = 5$ has an initial $time_{rem} = 125$. All averages are computed across 10 trials.

\subsection{Comparison to Null Model}

Here we present three separate experiments which compare the dRAP and FIFO algorithms. The first is a simple timing comparison that looks at $T_{complete}$ and $T_{wait}$ for each case. Values are presented in Table \ref{tab:timing} (including 95\% confidence intervals).

\begin{table}
\begin{center}
\begin{tabular}{|c|c|c|}
\hline
 & $T_{complete}$ & $T_{wait}$ \\
 \hline
dRAP & 845.60 (861.94,829.26) & 342.54 (349.30,335.78) \\
FIFO & 1071.20 (1088.99,1053.41) & 475.31 (485.79,464.82)\\
\hline
\end{tabular}
\caption{Average timing comparison of dRAP and FIFO scheduling algorithms with 95\% confidence intervals.}
\label{tab:timing}
\end{center}
\end{table}

We observe an approximate 20\% reduction in $T_{complete}$ and an approximate 25\% reduction in $T_{wait}$ when comparing dRAP to FIFO.

Our second experiment comparing dRAP and FIFO involves average cluster utilization. Because dRAP assigns tasks s.t. $CPU_{cluster} == CPU_{req}$, this ensures that all nodes in the cluster will be utilized. However, the FIFO scheduling system hands out tasks to the first available cluster, meaning it allows for the possibility that $CPU_{cluster} > CPU_{req}$. For example, a task with $CPU_{req} = 2$ that is assigned to a cluster with $CPU_{cluster} = 5$ will leave 3 unused nodes. Thus, we present an analysis of cluster utilization using the metric in Equation \ref{eq:mu}:

\begin{equation} \label{eq:mu} \mu_{cluster} = \frac{CPU_{req}}{CPU_{cluster}} \end{equation}

If $CPU_{cluster} \le CPU_{req}$, we simply set $\mu_{cluster} = 1$. Values are presented as percentages in Table \ref{tab:util} (note that dRAP's $\mu_{cluster}$ is always 100\% by definition).

\begin{table}
\begin{center}
\begin{tabular}{|c|c|}
\hline
 & $\mu_{cluster}$ \\
 \hline
dRAP & 100\% \\
FIFO & 56\% (54\%,58\%)\\
\hline
\end{tabular}
\caption{Average cluster utilization of dRAP and FIFO scheduling algorithms with 95\% confidence intervals}
\label{tab:util}
\end{center}
\end{table}

\begin{figure}
\begin{center}
	\scalebox{0.6}{\includegraphics{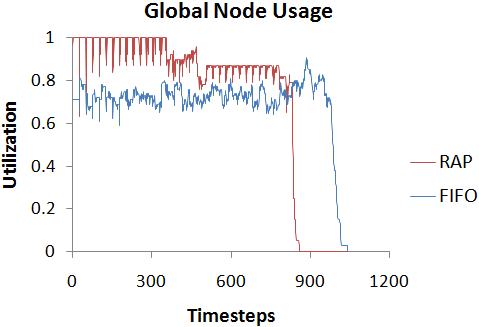}}
	\caption{Global utilization of nodes throughout simulation}
	\label{fig:util}
\end{center}
\end{figure}

Finally, our third experiment is designed to measure global node utilization over the time of the simulation. Here we simply document the number of nodes that do computation on a given timestep and normalize by the total number of nodes in the system. Results are displayed in Figure \ref{fig:util} (taken from a single simulation run).

We observe that the dRAP algorithm utilizes approximately 90-95\% of the nodes for the majority of the simulation, while FIFO utilizes approximately 70-75\%.

\subsection{Effective Complexity}

\begin{figure}
\begin{center}
	\scalebox{0.6}{\includegraphics{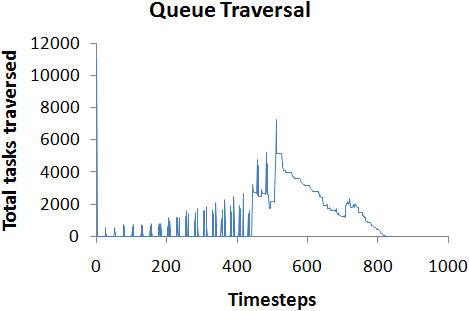}}
	\caption{Total task traversals by all clusters per timestep.}
	\label{fig:qtraversal}
\end{center}
\end{figure}

For this experiment, we estimate the ``effective'' computational complexity of the dRAP algorithm. That is, in comparison to the qualitative $O(nm)$ worst case runtime per timestep, we are interested in how much of the task queue must be traversed in order to properly fit the $CPU_{cluster} == CPU_{req}$ requirement. Total tasks traversed per timestep from one selected simulation run are presented in Figure \ref{fig:qtraversal}. Note that the initial traversal (timestep ``0''), although difficult to see, is approximately 11,000.

``Worse case'' here, as addressed above, is $O(nm)$, or $100,000$ tasks traversed per timestep if $n$ = number of clusters = number of nodes = 100 and $m$ = number of tasks = 1000. From this plot (plus additional runs not included here), we can conclude that effective computational complexity is no more than approximately 10\% of the worst case runtime $O(nm)$.

\subsection{Scaling}

\begin{figure}
\begin{minipage}[c]{0.4\linewidth}
\centering
\includegraphics[scale=.5]{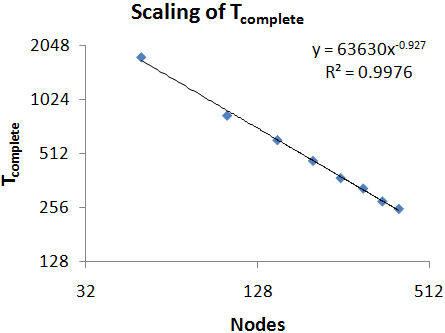}
\caption{Scaling of $T_{complete}$}
\label{fig:scalingtcomplete}
\end{minipage}
\hspace{0.8cm}
\begin{minipage}[c]{0.4\linewidth}
\centering
\includegraphics[scale=.5]{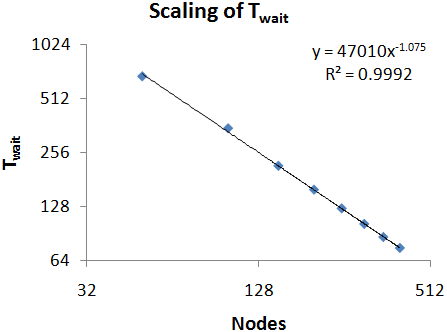}
\caption{Scaling of $T_{wait}$}
\label{fig:scalingtwait}
\end{minipage}
\end{figure}

For our last experiment, we are interested in collecting information on the scaling ability of our algorithm. Our goal in this test is to increase the number of nodes (in intervals of 50), while also maintaining an equal number of neighbors for each node. That is, we ensure that the neighborhood size parameter defined in the simulation scales inversely with the number of nodes s.t. a given node has approximately the same number of neighbors regardless of the total nodes in the system. Results are presented for our two timing metrics: $T_{complete}$ scaling in Figure \ref{fig:scalingtcomplete} and $T_{wait}$ scaling in Figure \ref{fig:scalingtwait}. Data in both figures are log$_2$-transformed in order to correlate doubling of nodes with halving of the timing metrics.

\vspace{5 mm}

We note a near-perfect scaling for both timing metrics, as shown in the fitted power law equations inset into each plot (Figures \ref{fig:scalingtcomplete} and \ref{fig:scalingtwait}). Note that the $T_{wait}$ exponent above 1 is most likely a result of inexact tuning of the neighborhood size with increasing nodes, and this issue will rectified in future work.

\vspace{5 mm}

\section{Related Work}

Resource allocation for grid computing is an active area of research. For example, SLURM \cite{yoo2003slurm} is a configurable Linux utility for cluster allocation that uses static allocation of nodes to clusters, called partitions, in contrast to the dynamic cluster allocation presented in this paper. LSF \cite{zhou1992lsf} is another proprietary cluster management facility, however details of its allocation algorithm are not publicly available.

\section{Conclusions and Future Work}

In this paper we have presented an algorithm for allocating, scheduling and load-balancing processes on a massively distributed system. This is very relevant to current research in operating systems, especially with a trend of moving computation tasks onto inexpensive, distributed hardware. The proposed decentralized algorithm draws inspiration from biology, adaptively creating and dissociating clusters from nodes to match task demand. Decentralization enables scalability, robustness, alleviation of computing load on monitor, better response and adaptability to process queue fluctuations and learning about process requirements. The dRAP algorithm outperforms a FIFO scheduling algorithm on time to complete all tasks, average waiting time and CPU utilization. The scheduling is also shown to be robust to a malicious adversary that might permute the order of the tasks such that high demand tasks would be queued first followed by low demand tasks. 
A key feature of our procedure is that nodes communicate with neighboring computers in order to dynamically form clusters. Hence our algorithm also holds promise in areas where it is advantageous to communicate with immediate neighbors due to network latency, e.g. Google MapReduce uses a locality optimization to reduce latency due to network communication \cite{dean2004mapreduce}. The comparison of this algorithm to other scheduling algorithms like SRTF (Shortest Remaining Time First) on other metrics like response time, as well as collection of data on the exact distribution of process demand in a queue in a real-world scenario, will be the subject of future investigation.

\section{Acknowledgements}
We thank Dr. Dorian Arnold for fruitful discussions.
\bibliographystyle{splncs03}
\bibliography{bib}

\begin{thebibliography}{10}
\providecommand{\url}[1]{\texttt{#1}}
\providecommand{\urlprefix}{URL }

\bibitem{anderson2002seti}
Anderson, D., Cobb, J., Korpela, E., Lebofsky, M., Werthimer, D.: {SETI@ home:
  an experiment in public-resource computing}. Communications of the ACM
  45(11), ~61 (2002)

\bibitem{Banerjee2009a}
Banerjee, S.: {An Immune System Inspired Approach to Automated Program
  Verification}. arXiv preprint arXiv:0905.2649  (2009),
  \url{http://arxiv.org/abs/0905.2649}

\bibitem{banerjee_phd}
Banerjee, S.: {Scaling in the immune system (PhD Thesis, University of New
  Mexico)} (2013)

\bibitem{Banerjee2011}
Banerjee, S., Levin, D., Moses, M., Koster, F., Forrest, S.: {The Value of
  Inflammatory Signals in Adaptive Immune Responses}. In: Artificial Immune
  Systems. pp. 1--14. Springer (2011),
  \url{http://www.springerlink.com/index/U634HJ83W62W5383.pdf}

\bibitem{Banerjee2009}
Banerjee, S., Moses, M.: {A Hybrid Agent Based and Differential Equation Model
  of Body Size Effects on Pathogen Replication and Immune System Response}. In:
  Timmis, J. (ed.) The 8th International Conference on Artificial Immune
  Systems (ICARIS). pp. 14--18. Springer, Lecture Notes in Computer Science
  (2009), \url{http://www.springerlink.com/content/b786g874642q2j37/}

\bibitem{banerjee2010modular}
Banerjee, S., Moses, M.: Modular radar: An immune system inspired search and
  response strategy for distributed systems. In: Artificial Immune Systems, pp.
  116--129. Springer (2010)

\bibitem{swarm2010}
Banerjee, S., Moses, M.: {Scale Invariance of Immune System Response Rates and
  Times: Perspectives on Immune System Architecture and Implications for
  Artificial Immune Systems}. Swarm Intelligence  4(4),  301--318 (2010),
  \url{http://www.springerlink.com/content/w67714j72448633l/}

\bibitem{cook2004universality}
Cook, M.: {Universality in elementary cellular automata}. Complex Systems
  15(1),  1--40 (2004)

\bibitem{dean2004mapreduce}
Dean, J., Ghemawat, S.: {MapReduce: Simplified Data Processing on Large
  Clusters}. To appear in OSDI p.~1 (2004)

\bibitem{gardner1970mathematical}
Gardner, M.: {Mathematical games: The fantastic combinations of John Conway's
  new solitaire game 'Life'}. Scientific American  223(4),  120--123 (1970)

\bibitem{Hecker2012}
Hecker, J.P., Letendre, K., Stolleis, K., Washington, D., Moses, M.E.: {Formica
  ex machina: Ant swarm foraging from physical to virtual and back again}. In:
  Swarm Intelligence: 8th International Conference, ANTS 2012, pp. 252--259.
  Springer Berlin Heidelberg, Berlin, DE (2012)

\bibitem{Hecker2013}
Hecker, J.P., Moses, M.E.: {An evolutionary approach for robust adaptation of
  robot behavior to sensor error}. In: Proceedings of the 15th Annual
  Conference Companion on Genetic and Evolutionary Computation (GECCO '13
  Companion). pp. 1437--1444. ACM, New York, NY (2013),
  \url{http://doi.acm.org/10.1145/2464576.2482724}

\bibitem{Hecker2015}
Hecker, J.P., Moses, M.E.: {Beyond pheromones: Evolving error-tolerant,
  flexible, and scalable ant-inspired robot swarms}. Swarm Intelligence  9(1),
  43--70 (2015)

\bibitem{Hecker2013a}
Hecker, J.P., Stolleis, K., Swenson, B., Letendre, K., Moses, M.E.: {Evolving
  error tolerance in biologically-inspired iAnt robots}. In: Proceedings of the
  Twelfth European Conference on the Synthesis and Simulation of Living Systems
  (Advances in Artificial Life, ECAL 2013). pp. 1025--1032. MIT Press,
  Cambridge, MA (2013)

\bibitem{hecker2008team}
Hecker, J., Wu, A., Herweg, J., Sciortino~Jr, J.: {Team-based resource
  allocation using a decentralized social decision-making paradigm}. In:
  Proceedings of SPIE. vol. 6964, p. 696409 (2008)

\bibitem{jacob2004immunity}
Jacob, C., Litorco, J., Lee, L.: {Immunity through swarms: Agent-based
  simulations of the human immune system}. Lecture Notes in Computer Science
  pp. 400--412 (2004)

\bibitem{krawczyk2008grid}
Krawczyk, S., Bubendorfer, K.: {Grid resource allocation: allocation mechanisms
  and utilisation patterns}. In: Proceedings of the sixth Australasian workshop
  on Grid computing and e-research-Volume 82. pp. 73--81. Australian Computer
  Society, Inc. (2008)

\bibitem{luke2004mason}
Luke, S., Cioffi-Revilla, C., Panait, L., Sullivan, K.: {Mason: A new
  multi-agent simulation toolkit}. In: Proceedings of the 2004 SwarmFest
  Workshop. vol.~8 (2004)

\bibitem{alife_radar}
Moses, M., Banerjee, S.: Biologically inspired design principles for scalable,
  robust, adaptive, decentralized search and automated response (radar). In:
  Artificial Life (ALIFE), 2011 IEEE Symposium on. pp. 30--37 (april 2011)

\bibitem{yoo2003slurm}
Yoo, A., Jette, M., Grondona, M.: {SLURM: Simple linux utility for resource
  management}. Lecture Notes in Computer Science pp. 44--60 (2003)

\bibitem{zhou1992lsf}
Zhou, S.: {LSF: load sharing in large-scale heterogeneous distributed systems}.
  In: Proc. Workshop on Cluster Computing. pp. 1995--1996 (1992)

\end{thebibliography}



\end{document}